\documentclass[conference]{IEEEtran}
\IEEEoverridecommandlockouts
\usepackage{cite}
\usepackage{amsmath,amssymb,amsfonts}
\usepackage{algorithmic}
\usepackage{graphicx}
\usepackage{textcomp}
\usepackage{xcolor}
\usepackage{booktabs}

\usepackage{bm}
\usepackage{subcaption}
\usepackage{xspace}
\usepackage{enumitem}
\newcommand{\mname}{\texttt{CCS Explorer}\xspace}
\newcommand{\pio}{\textit{PIO Detector}\xspace}
\newcommand{\summarizer}{\textit{Summarizer}\xspace}
\newcommand{\relevance}{\textit{Relevance Predictor}\xspace}
\newcommand{\query}{\textit{Query Generator}\xspace}
\newcommand{\gui}{\textit{Web Application}\xspace}
\newcommand{\artext}{\textit{Article Extractor}\xspace}
\usepackage{float}

\makeatletter
\newcommand{\linebreakand}{%
  \end{@IEEEauthorhalign}
  \hfill\mbox{}\par
  \mbox{}\hfill\begin{@IEEEauthorhalign}
}
\makeatother

\usepackage{multirow}
\def\BibTeX{{\rm B\kern-.05em{\sc i\kern-.025em b}\kern-.08em
    T\kern-.1667em\lower.7ex\hbox{E}\kern-.125emX}}
\begin{document}

\title{\mname: Relevance Prediction, Extractive Summarization, and Named Entity Recognition from Clinical Cohort Studies\\
% \thanks{We would like to thank the wonderful team at Morningside Center for Innovative and Affordable Medicine. This research was funded by National Science Foundation CAREER grant 1944247 to C.M, National Institute of Health grant U19-AG056169 sub-award to C.M., and the McCamish Parkinson’s Disease Innovation Program at Georgia Institute of Technology and Emory University to C.M.}
}

\author{\IEEEauthorblockN{Irfan Al-Hussaini}
\IEEEauthorblockA{\textit{Electrical and Computer Engineering} \\
\textit{Georgia Institute of Technology}\\
Atlanta, GA, USA \\
alhussaini.irfan@gatech.edu}
\and
\IEEEauthorblockN{Davi Nakajima An}
\IEEEauthorblockA{\textit{Computer Science} \\
\textit{Georgia Institute of Technology}\\
Atlanta, GA, USA \\
dna@gatech.edu}
\and
\IEEEauthorblockN{Albert J. Lee}
\IEEEauthorblockA{\textit{Machine Learning} \\
\textit{Georgia Institute of Technology}\\
Atlanta, GA, USA \\
albert.jb.lee@gatech.edu}

\linebreakand

\IEEEauthorblockN{Sarah Bi}
\IEEEauthorblockA{\textit{Biomedical Engineering} \\
\textit{Georgia Institute of Technology}\\
Atlanta, GA, USA \\
sbi30@gatech.edu}
\and
\IEEEauthorblockN{Cassie S. Mitchell}
\IEEEauthorblockA{\textit{Biomedical Engineering and Machine Learning} \\
\textit{Georgia Institute of Technology and Emory University}\\
Atlanta, GA, USA \\
cassie.mitchell@bme.gatech.edu}
}

\maketitle

\begin{abstract}
Clinical Cohort Studies (CCS), such as randomized clinical trials, are a great source of documented clinical research. Ideally, a clinical expert inspects these articles for exploratory analysis ranging from drug discovery for evaluating the efficacy of existing drugs in tackling emerging diseases to the first test of newly developed drugs. However, more than 100 articles are published daily on a single prevalent disease like COVID-19 in PubMed. As a result, it can take days for a physician to find articles and extract relevant information. Can we develop a system to sift through the long list of these articles faster and document the crucial takeaways from each of these articles? In this work, we propose \mname, an end-to-end system for relevance prediction of sentences, extractive summarization, and patient, outcome, and intervention entity detection from CCS. \mname is packaged in a web-based graphical user interface where the user can provide any disease name. \mname then extracts and aggregates all relevant information from articles on PubMed based on the results of an automatically generated query produced on the back-end. For each task, \mname fine-tunes pre-trained language representation models based on transformers with additional layers. The models are evaluated using two publicly available datasets. \mname obtains a recall of 80.2\%, AUC-ROC of 0.843, and an accuracy of 88.3\% on sentence relevance prediction using BioBERT and achieves an average Micro F1-Score of 77.8\% on Patient, Intervention, Outcome detection (PIO) using PubMedBERT. Thus, \mname can reliably extract relevant information to summarize articles, saving time by $\sim \text{660}\bm{\times}$.

\end{abstract}


\section{Introduction}
One of the world’s largest biomedical publication databases, PubMed, has over 34 million publications. Approximately 2.5 million users perform about 3 million searches and 9 million page views on PubMed every day \cite{white2020pubmed}. Over the past couple of years, 137 articles have been posted per day on PubMed on COVID-19 alone \cite{yeo2021alarming}. In particular, clinical Cohort Studies (CCS), which contain information on the specific results of a patient or patient population for a given therapeutic and/or condition, are considered essential for clinical research. Clinical cohort studies include randomized clinical trials, prospective cohort studies, retrospective cohort studies, case-control studies, patient case studies, and more. Clinical cohort studies typically describe a patient or patient population, the intervention(s) assessed, and the measured outcome(s).

\begin{figure}[t]\framebox[\linewidth]{
    \centering
    \captionsetup{justification=centering}
    \includegraphics[width=0.99\linewidth]{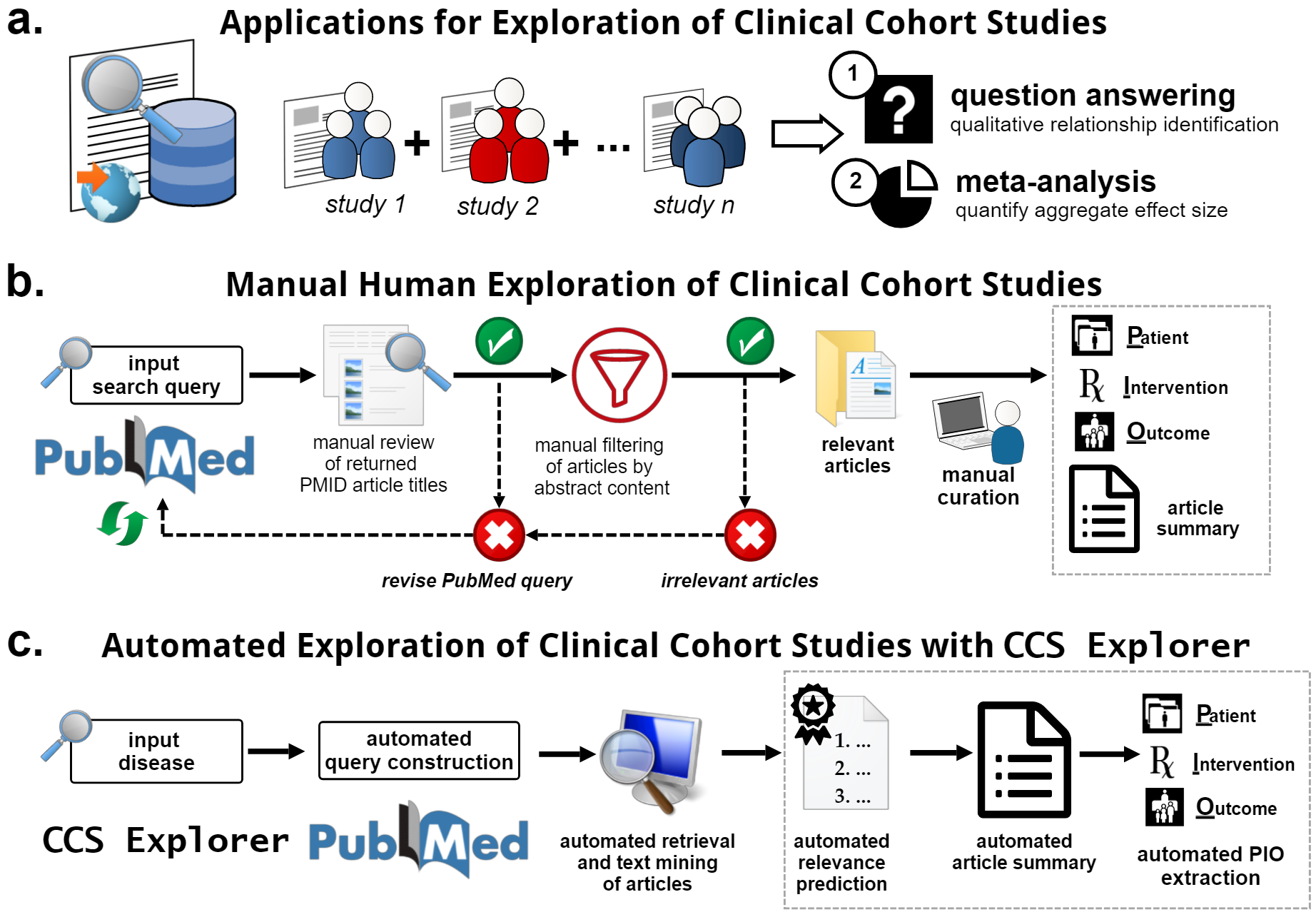}}
    \caption{Why \mname?}
    \label{fig:compare_manual}
\end{figure}

The two major applications that require the exploration of clinical cohort studies are question-answering and meta-analysis (Figure 1a). Investigation of clinical cohort studies is necessary to answer questions and identify qualitative relationships. Examples of question answering include: What drugs may be repurposed or used in combination to improve disease outcomes \cite{mccoy2021biomedical}? What comorbidities are most impactful to cardiac disease outcome \cite{burke2007interpretation, cavailles2013comorbidities, listerman2011cardiac, lang2007non}? What patient features result in health outcome disparities \cite{arnold2008age, ashwell2012waist}? Exploration of clinical cohort studies is also required to perform a meta-analysis, a quantitative analysis where the results of cohort studies are aggregated to estimate an overall effect size. Estimating an overall or aggregate effect size, such as the effect of a drug on disease outcome, adjusts for disparity or bias introduced by individual study-specific features (e.g., geography, gender, age, sample size). Examples of meta-analysis include: determining overall adverse event rates with specific treatments for cancer \cite{mohanavelu2021meta}, determining the overall prevalence of comorbidities in a rare neurodegenerative disease population \cite{mitchell2015antecedent}, or determining the overall effect size of vaccination on SARS-CoV-2 outcome \cite{makhoul2020epidemiological, pritchard2021impact, shattock2022impact}.

The process for manual exploration of cohort studies is iterative and time-consuming (Figure 1b). The major steps include devising the appropriate advanced PubMed query to find articles in PubMed, reviewing the list of search title results to determine if the query resulted in the expected type or number of studies, examining the abstracts to determine if the journal article contains the desired information, and curating the article to extract the pivotal PIO elements: patient population (disease and/or control population), intervention (what therapy was utilized), and the outcome (what measurement was utilized to determine a result). Depending on the number of studies to be reviewed and included, the exploration process alone can take hours to weeks before final curation and analysis can occur \cite{mitchell2015undergraduate}. Moreover, even with a quality control team, there may be some remaining inconsistency between researchers or curators \cite{mitchell2015undergraduate}. Critical variations and corresponding delays may occur depending on the researcher’s knowledge of constructing an appropriate advanced PubMed query. An appropriate PubMed query must include all relevant synonyms, MESH terms, and proper formatting to return the most inclusive and relevant list of articles. Additionally, differences in review styles for examining lengthy abstracts or even full-text articles may result in unintended differences in article inclusion or stylistic differences in PIO extraction.

Although there are specially trained groups dedicated to manually synthesizing findings from CCS, the rapid publication of new articles makes it impossible to maintain pace \cite{tsafnat2013automation}. However, Natural Language Processing (NLP) breakthroughs have enabled the automation of many time-consuming tasks related to text exploration in non-biomedical domains such as sentiment analysis of customer reviews \cite{sentiment}, language translation \cite{translation1, translation2}, ranking search results \cite{anyanwu2005semrank, belter2017relevance, gao2020toward}, abstractive summarization \cite{zhang2020pegasus, maynez2020faithfulness, gehrmann2018bottom}, and extractive summarization \cite{zhong2020extractive, liu2019fine}. %The presented work applies and integrates NLP tasks to automate biomedical text review, namely clinical cohort study exploration.

Here we present \mname to automate the clinical cohort study exploration (Figure 1c). \mname is an open-source web application that dramatically expedites identifying, reviewing, and extracting data from clinical cohort studies. CCS only requires that the user input a disease name. Using a pre-built list of intervention names (which can also be customized if desired), \mname formulates an advanced query to PubMed.  \mname automatically obtains all relevant articles via their unique PubMed identification (PMID) and parses through the text.  \mname provides three critical outputs for researchers: 1) a list of all relevant studies along with a relevance prediction score; 2) an abbreviated relevance summary that contains only the most relevant information (or sentences) necessary for the researcher to explore the study; 3) automated extraction of PIO elements. With \mname, question answering or meta-analysis is greatly expedited, streamlined, and optimized. \mname automates all the iterative, front-end work that generally takes a specially trained team of researchers hours to weeks to achieve.

%When exploring CCS, there is a discrete sequence of steps the curator must take as shown in Figure 1b: (1) create a query on PubMed to find relevant articles; (2) find relevant entities in the article like cohort demography; (3) write a summary of the paper with key takeaways. Can we build an end-to-end system to automate this sequence of tasks? 

\begin{figure}[t]
    \centering
    \captionsetup{justification=centering}
    \includegraphics[width=\linewidth]{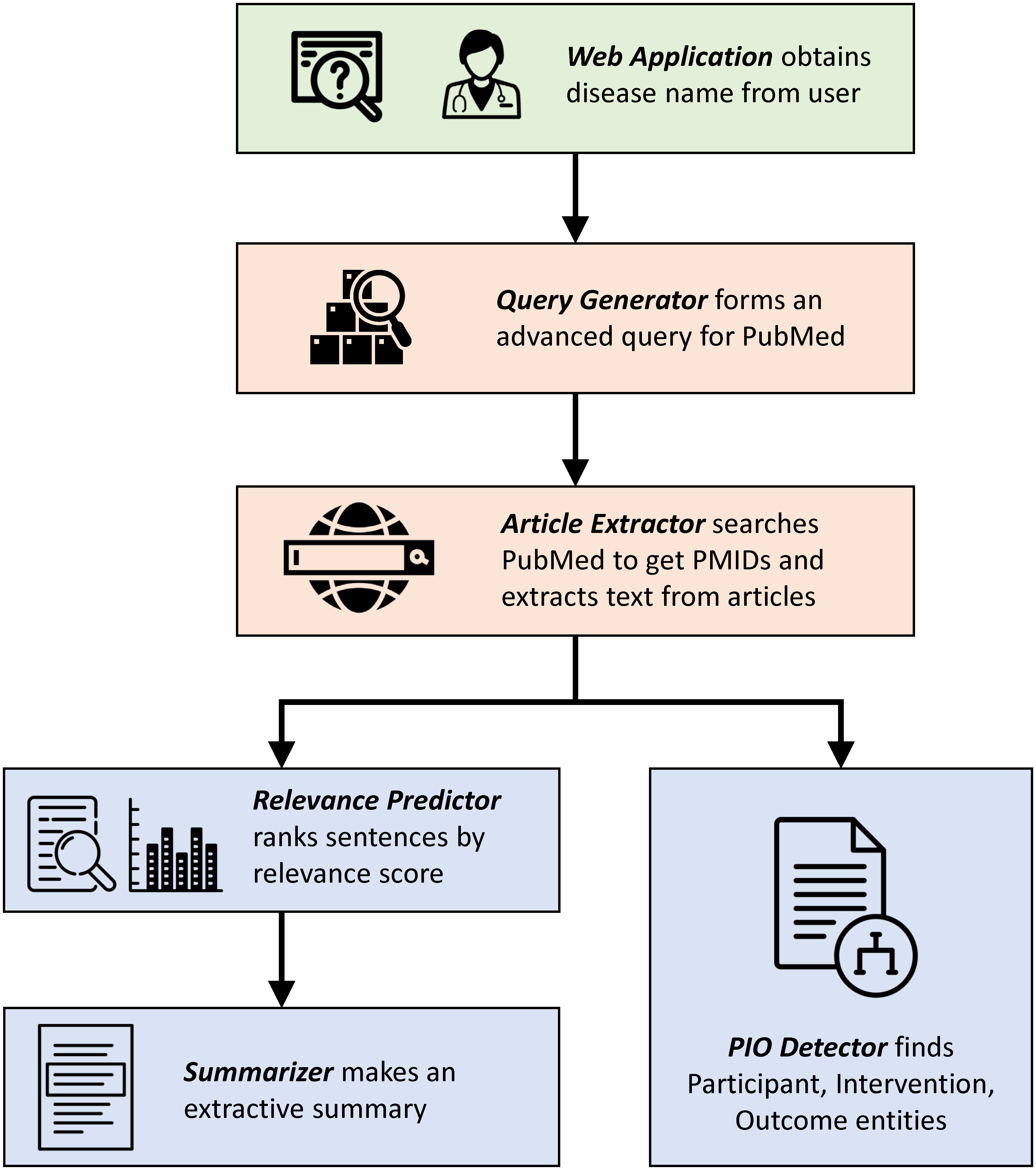}
    \caption{\mname Framework}
    \label{fig:framework}
\end{figure}

\mname is an end-to-end system for exploring clinical cohort studies with PubMed and extracting useful information necessary for tasks like question answering or meta-analysis. Figure \ref{fig:compare_manual} highlights the difference between manual exploration by an expert and \mname. It can reduce the time taken to extract relevant information and summarize articles from hours to seconds. For the query demonstrated in Figure \ref{fig:gui}, it takes 26.32s for \mname to run the query, process the resulting articles, and extract all relevant information to construct a task-specific summary along with the detection of PIO entities. 

We introduce two models for this task, one for relevance prediction of text and another for detecting participant, intervention, and outcome (PIO) entities in CCS articles. We compare the proposed models' performance by initializing the weights using 6-7 different pre-trained BERT \cite{bert}, and ELECTRA \cite{clark2020electra} models.

The main contribution of this paper is to design each of the following pieces and combine them to form \mname:
\begin{itemize}[leftmargin=*]
    \item \gui: front-end designed for taking inputs from the user and displaying outputs
    \item \query: merges MeSH terms to form an advanced query for PubMed to obtain PMIDs
    \item \artext: extracts articles from PubMed and stores the text for subsequent steps
    \item \relevance: attention-based language model that assigns a relevance score to each sentence in the article
    \item \summarizer: generates an extractive summary of the article by putting together the most relevant sentences into a coherent body of text
    \item \pio: named entity recognition model finds participants, interventions, and outcomes present in the article and assigns a  score for each entity
\end{itemize}
The resulting framework of \mname is shown in Figure \ref{fig:framework}.

\section{System Design}

\begin{figure}\framebox[\linewidth]{
\begin{minipage}{\linewidth}
     \centering
     \begin{subfigure}[t]{0.99\linewidth}
         \centering
         \captionsetup{justification=centering}
         \includegraphics[width=\linewidth]{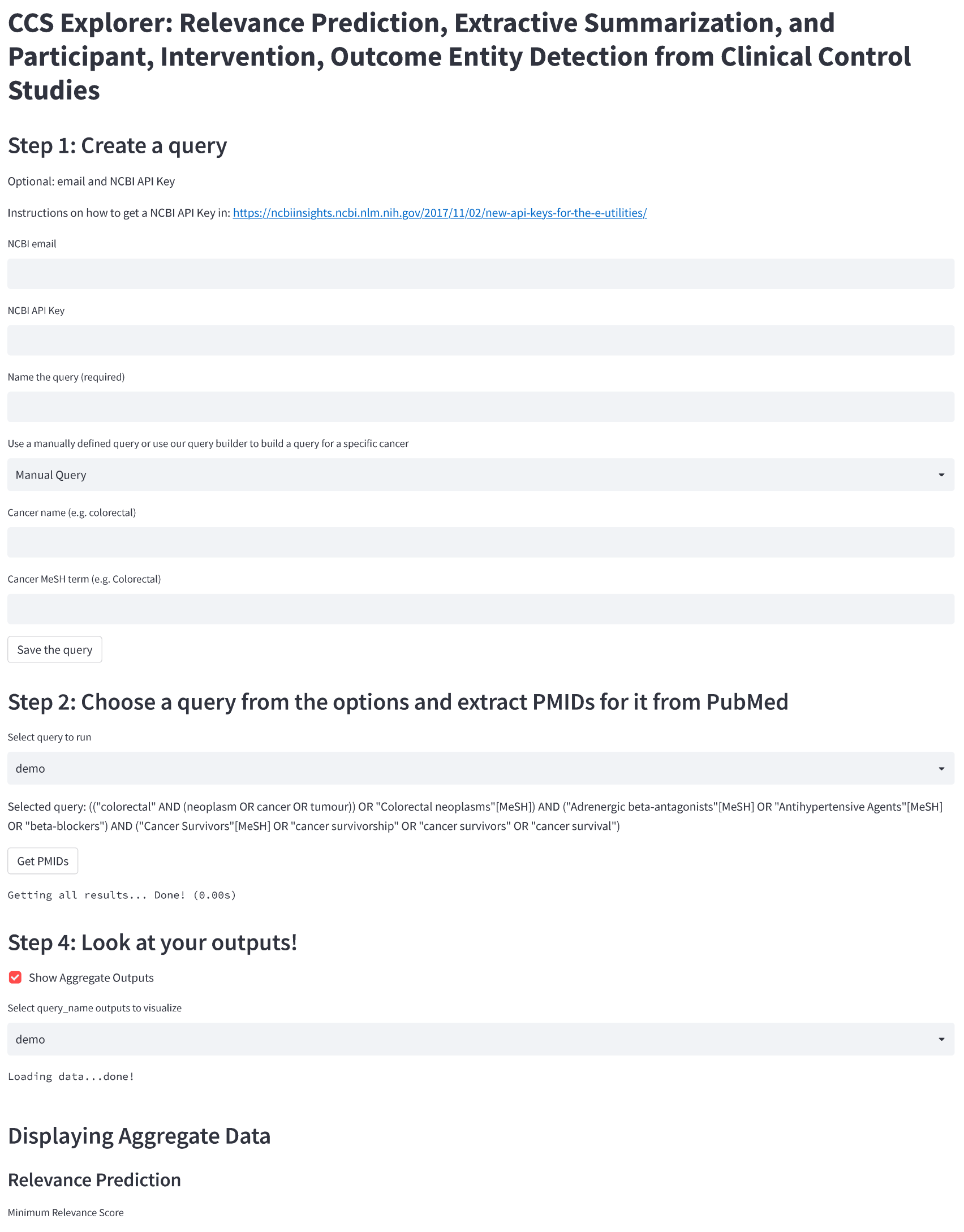}
        %  \caption{t-distributed stochastic neighbor embedding (t-SNE)}
        %  \label{fig:tsne}
     \end{subfigure}
    %  \hfill
     \begin{subfigure}[t]{0.99\linewidth}
         \centering
         \captionsetup{justification=centering}
         \includegraphics[width=\linewidth]{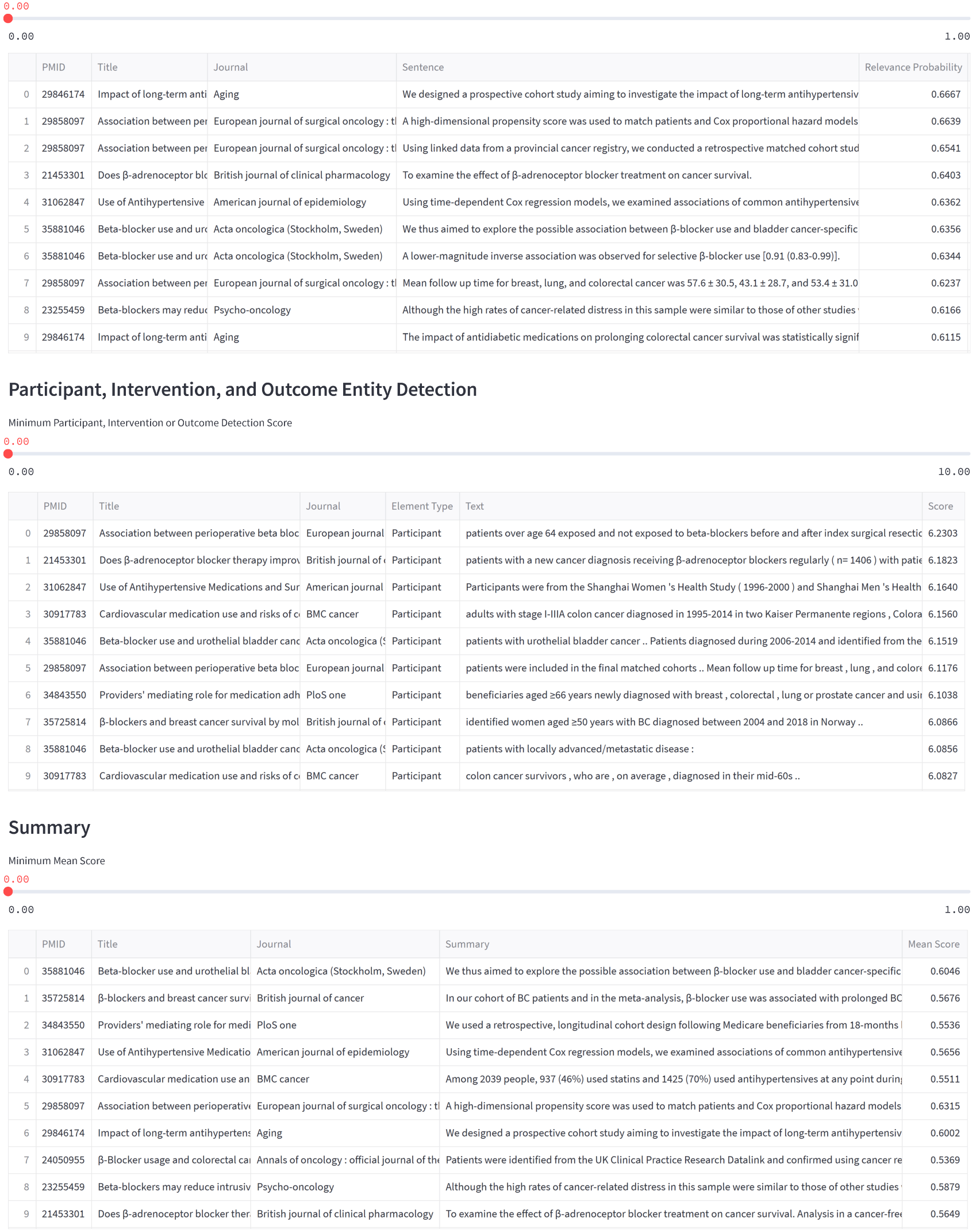}
        %  \caption{Principal component analysis\\(PCA)}
        %  \label{fig:pca}
     \end{subfigure}\end{minipage}}
        \caption{\mname: Graphical User Interface}
        \label{fig:gui}
\end{figure}

The goal of \mname is to provide a user-friendly system for researchers to obtain reliable results expeditiously. To this end, Streamlit \cite{streamlit} was used to design an intuitive front-end user interface for \mname. The graphical user interface, GUI, is shown in Figure \ref{fig:gui}. It takes inputs from the user and displays the results in a user-friendly format for review. \mname comprises of the following: 
\begin{itemize}[leftmargin=*]
    \item Step 1: This encompasses the creation of the query. The user has two options: (1) create a manual query by stitching together MeSH terms (2) provide a disease name so that the \query can build an advanced query for PubMed. The query is named to enable usage in subsequent steps.
    \item Step 2: A query name has to be selected from the options provided consisting of previously formed queries. PMIDs are obtained from PubMed based on the selected query.
    \item Step 3: The articles are extracted from PubMed using the PMIDs, and \relevance, \summarizer, and \pio are run on each article to obtain aggregated results. This step is not visible in Figure \ref{fig:gui} since it is complete, and the user has moved on to the next step.
    \item Step 4: Three tables show the results of \relevance, \summarizer, and \pio.
\end{itemize}
% \vspace{0.5em}
\noindent \mname can be divided into three different pieces which run in the back-end: (1) \query and \artext, (2) \relevance and \summarizer, (3) \pio. The details of \relevance are discussed in Section \ref{sec:relevance} and \pio in Section \ref{sec:pio}. The framework of \mname is shown in Figure \ref{fig:framework}.

\vspace{0.5em}
\noindent \textbf{Query Generator and Article Extraction.}
\mname provides the users with a graphical user interface to input their National Center for Biotechnology Information (NCBI) email and API key to enable repeated PubMed queries. It also allows the user to manually input a customized advanced query using MeSH terms or to provide a cancer type so that an automatically generated query can obtain a baseline result. The query generation and extraction of articles are performed using BioPython \cite{biopython_1, biopython_2}. Each of the resulting texts is prepared for subsequent steps using SciSpacy \cite{scispacy}.

\section{Relevance Prediction and Extractive Summarization}
\label{sec:relevance}
\subsection{Data}
\label{sec:rel_data}
The data used in building the \relevance and \summarizer of \mname originate from an open source dataset called the Evidence Inference dataset \cite{relevance_data1, relevance_data2, relevance_model}. It contains valuable annotations of relevant information in CCS articles.

In this dataset, \cite{relevance_data1, relevance_data2, relevance_model}, groups of text are labeled as \textit{evidence} and \textit{nonevidence}. In designing \mname, we replaced the \textit{evidence} label with \textit{relevant} and \textit{non-evidence} with \textit{irrelevant}. The resulting \textit{relevant} and \textit{irrelevant} labels were used as ground truth annotations to build the \relevance of \mname. It consists of 4,005 unique articles split across two partitioned sets. The selection of articles for training and test set was defined in the Evidence Inference \cite{relevance_data1, relevance_data2} dataset as train\textunderscore article\textunderscore ids and validation\textunderscore article\textunderscore ids, respectively.

The \summarizer uses results from the \relevance to formulate summaries and a Summary Score to denote its quality.

\subsection{Method}
\label{sec:relevance_method}

\noindent \textbf{Relevance Prediction.} \relevance was designed using BERT-based language models pre-trained on scientific articles obtained from sources such as PubMed, PubMed Central, and UMLS. It was constructed by adding a dense layer to the pre-trained model architecture and fine-tuned on the Evidence Inference dataset described in Section \ref{sec:rel_data}. 

The pre-trained BERT models used: 
\begin{itemize}[leftmargin=*]
    \item BioBERT \cite{biobert}: Initialized using standard BERT \cite{bert} model, and then pre-trained on Biomedical domain texts, which includes PubMed abstracts and PubMed Central full-text articles.
    \item PubMedBERT \cite{pubmedbert}: Pretrained a BERT \cite{bert}  model from scratch using 14 million abstracts from PubMed. 
    \item SapBERT \cite{sapbert}: Pre-trained a BERT model on the biomedical knowledge graph of UMLS \cite{umls} using self-alignment to cluster synonyms of the same concept.
    \item BlueBERT \cite{bluebert}: Initialized using standard BERT \cite{bert} model and pre-trained on PubMed abstracts (4 Billion words) and clinical notes from MIMIC-III (500 Million words). \cite{johnson2016mimic}.
    \item KRISSBERT \cite{krissbert}: Initialized with PubMedBERT \cite{pubmedbert} parameters, and then pre-trained using biomedical entity names from the UMLS ontology \cite{umls} to self-supervise entity linking examples from PubMed abstracts.
    \item SciBERT  \cite{scibert}: Trained a BERT \cite{bert} model on scientific papers taken from 1.14 million full papers from Semantic Scholar.
\end{itemize}

Let $\mathcal{Y^\prime}$ be all the outputs from the model, $\mathcal{Y}$ be all the annotations from the dataset, $y^\prime_i \in [0,1]$ represent the model prediction, and $\bm{y}_i$ denote the annotation of the $i$-th sentence. Let $\bm{h}(\mathcal{X})$ represent the output of the transformer architecture. This output, $\bm{h}(\mathcal{X})$, is used as input to a fully-connected layer followed by the sigmoid function ($\sigma$). So, the output of the model for the $i$-th sentence is represented by: %$\bm{z}_i = \bm{W}^\top \bm{h}(\bm{x}_i) + \bm{b}$. 

\begin{equation}\label{eq:rel_model}
\begin{aligned}
    \bm{z}_i &= \bm{W}^\top \bm{h}(\bm{x}_i) + \bm{b} \\
    \bm{y^\prime}_i &= \sigma(\bm{z}_i) =  \frac{1}{1+e^{-\bm{z}_i}}
\end{aligned}
\end{equation}

% Let $\mathcal{Y^\prime}$ be all the outputs from the model and $\mathcal{Y}$ be all the annotations from the dataset, $y^\prime_i \in [0,1]$ represent the model prediction and annotation of the $i$-th token.

Binary cross entropy loss is used and is denoted by:
\begin{equation}\label{eq:rel_loss}
\begin{aligned}
% L = l(\mathcal{Y^\prime},\mathcal{Y}) = \{l_1, \dots ,l_N\}^T \\
L(\bm{y}_i,\; \bm{y^\prime}_i) = &-[\bm{y}_i \cdot log(\bm{y}^\prime_i)\\
&+(1-\bm{y}_i) \cdot log(1-\bm{y}^\prime_i)]
\end{aligned}
\end{equation}

\vspace{0.5em}

\noindent \textbf{Summarization.} The output of the sigmoid function ($\sigma$) in Equation \eqref{eq:rel_model}, $\bm{y}^\prime_i$, represents the relevance score for the $i$-th sentence. The sentences are then sorted in descending order by these relevance scores to generate the set of sentences $\mathcal{Y}^\prime_{sorted}$. The first 4 sentences corresponding to the 4 most relevant sentences are joined to form the extractive summary for each article. The summary score is the average of the relevance scores for each of these 4 sentences. 
\begin{equation}
    \text{Summary Score} = \frac{\sum^4_{i=1} \bm{y}^\prime_{i,sorted}}{4}
\end{equation}
% Since PubMedBERT was determined as the best pretrained model for PIO extraction in terms of F1-Score, we decided to keep it as our base weights for Relevance Classification. We finetuned another 9 PubMedBERT model on the Evidence Inference dataset [23] (only using the ’evidence’ labels from the dataset) using a random subset of the original dataset with 20K sentences for finetuning and 2K sentences for validation.

\vspace{0.5em}
% \subsubsection{Experimental Settings}
\noindent\textbf{Metrics.} The following metrics were used to evaluate the performance of the relevance prediction model:
% \begin{itemize}[leftmargin=*]
\begin{equation}
\begin{aligned}
\text{Accuracy} &=\frac{\left|\mathcal{Y}\cap\mathcal{Y}^\prime\right|}{N} \\
\text{Recall}, R&=\frac{\left|\mathcal{Y}\cap\mathcal{Y}^{\prime}\right|}{\left|\mathcal{Y}\right|}\\
 \text{Precision},  P&=\frac{\left|\mathcal{Y}\cap\mathcal{Y}^{\prime}\right|}{\left|\mathcal{Y}^\prime\right|} \\
 \text{F1 score} &=\frac{2\ \ast\ P\ \ast\ R}{P\ +\ R\ }
\end{aligned}
\end{equation}
% \end{itemize}

where the annotated relevance labels of the entire dataset are denoted by $\mathcal{Y}$ and the model predictions by $\mathcal{Y^\prime}$; $\left|\mathcal{Y}\right|$ and $\left|{\mathcal{Y}^\prime}\right|$ represent the number of annotated tokens and the number of model predictions. In addition to the above metrics, the area under the receiver operating characteristics curve (AUC-ROC) is used for comparison.
\vspace{0.5em}

\noindent\textbf{Implementation Details.} We implemented \relevance using PyTorch \cite{pytorch,pytorch-lightning} and transformers \cite{huggingface-transformers}. The model was trained using a machine equipped with Intel Xeon Gold 6136 Processor, 376GB RAM, an Nvidia V100 GPU, and CUDA 11.4. During training, we used a batch size of 16, and a learning rate of $10^{-5}$. Each model was trained for 4 epochs using ADAM \cite{adam} as the optimization method. The 3,562 articles defined in train\textunderscore article\textunderscore ids are used as the training set, and the 443 articles specified in validation\textunderscore article\textunderscore ids list are used as the test set from the Evidence Inference Dataset \cite{relevance_data1, relevance_data2}.

\subsection{Result}
\label{sec:relevance_result}

\begin{table}[t]
\label{tab:relevance}
\caption{\mname: Relevance Prediction Model Performance}
\centering
\resizebox{\linewidth}{!}{%
\begin{tabular}{lccccc} \toprule
Model      & Accuracy & Precision & Recall & AUC-ROC & F1-Score \\ \midrule
BioBERT \cite{biobert}    & 0.883    & 0.083     & 0.802  & 0.843   & 0.150    \\
PubMedBERT \cite{pubmedbert} & 0.880    & 0.080     & 0.801  & 0.841   & 0.145    \\
SapBERT \cite{sapbert}   & 0.887    & 0.083     & 0.776  & 0.832   & 0.150    \\
BlueBERT \cite{bluebert}  & 0.875    & 0.078     & 0.817  & 0.846   & 0.143    \\
KRISSBERT \cite{krissbert} & 0.884    & 0.082     & 0.792  & 0.839   & 0.149    \\
SciBERT  \cite{scibert}  & 0.877    & 0.080     & 0.814  & 0.846   & 0.145  \\ \bottomrule 
\end{tabular}}
\end{table}

\begin{table*}[t]
\caption{\mname generated extractive summaries of the following query: ((\textquotesingle\textquotesingle colorectal\textquotesingle\textquotesingle\xspace  AND (neoplasm OR cancer OR tumour)) OR \textquotesingle\textquotesingle Colorectal neoplasms\textquotesingle\textquotesingle\xspace [MeSH]) AND (\textquotesingle\textquotesingle Adrenergic beta-antagonists\textquotesingle\textquotesingle\xspace [MeSH] OR \textquotesingle\textquotesingle Antihypertensive Agents\textquotesingle\textquotesingle\xspace [MeSH] OR \textquotesingle\textquotesingle beta-blockers\textquotesingle\textquotesingle\xspace ) AND (\textquotesingle\textquotesingle Cancer Survivors\textquotesingle\textquotesingle\xspace [MeSH] OR \textquotesingle\textquotesingle cancer survivorship\textquotesingle\textquotesingle\xspace  OR \textquotesingle\textquotesingle cancer survivors\textquotesingle\textquotesingle\xspace  OR \textquotesingle\textquotesingle cancer survival\textquotesingle\textquotesingle\xspace )}
\label{tab:summary}
\centering
\resizebox{\linewidth}{!}{%
\begin{tabular}{lllc}\toprule
PMID     & Title                                                                                                                                                                                 & Journal                                                                                                                                           & Summary Score \\ \midrule
24050955 \cite{24050955} & $\beta$-Blocker usage and colorectal cancer mortality: a nested case-control study in the UK Clinical Practice Research Datalink cohort.                                                    & Annals of oncology ...                                                                & 0.537         \\
35881046 \cite{35881046} & Beta-blocker use and urothelial bladder cancer survival: a Swedish register-based cohort study.                                                                                       & Acta oncologica (Stockholm, Sweden)                                                                                                               & 0.605         \\
29858097 \cite{29858097} & Association between perioperative beta blocker use and cancer survival following surgical resection.                                                                                  & European journal of surgical oncology ... & 0.631         \\
29846174 \cite{29846174}  & Impact of long-term antihypertensive and antidiabetic medications on the prognosis of post-surgical colorectal cancer: the Fujian ... & Aging                                                                                                                                             & 0.600         \\
34843550 \cite{34843550} & Providers' mediating role for medication adherence among cancer survivors.                                                                                                            & PloS one                                                                                                                                          & 0.554         \\
31062847 \cite{31062847} & Use of Antihypertensive Medications and Survival Rates for Breast, Colorectal, Lung, or Stomach Cancer.                                                                               & American journal of epidemiology                                                                                                                  & 0.566         \\
35725814 \cite{35725814} & $\beta$-blockers and breast cancer survival by molecular subtypes: a population-based cohort study and meta-analysis.                                                                       & British journal of cancer                                                                                                                         & 0.568        \\
23255459 \cite{pmid_summary_pico_23255459} & Beta-blockers may reduce intrusive thoughts in newly diagnosed cancer patients.                                                                                                       & Psycho-oncology                                                                                                                                   & 0.588         \\
30917783 \cite{30917783} & Cardiovascular medication use and risks of colon cancer recurrences and additional cancer events: a cohort study.                                                                     & BMC cancer                                                                                                                                        & 0.551         \\
21453301 \cite{21453301} & Does $\beta$-adrenoceptor blocker therapy improve cancer survival? Findings from a population-based retrospective cohort study.                                                             & British journal of clinical pharmacology                                                                                                          & 0.565    \\ \bottomrule
\end{tabular}}
\end{table*}

A high recall is essential for relevance prediction to detect all relevant sentences. It is acceptable for an automated system to include some irrelevant sentences as long as the significant ones appear at the top of the list. Prior research in machine translation show alignment with human expectation is highest when the optimization focuses on recall \cite{lavie2004significance}. User evaluation of interactive information retrieval performance \cite{su1994relevance} indicates recall is significantly more correlated with the users' expectation of success. Similarly, recall is more important than precision for downstream tasks such as summarization \cite{nenkova2006summarization}. Most of the evaluated models for relevance prediction displayed a recall above 80\%, AUC-ROC above 84\%, and accuracy above 88\%. The low F1-score is due to the low precision, which is less critical for tasks such as relevance prediction \cite{lavie2004significance, su1994relevance, nenkova2006summarization}. Due to the highest average metrics among all methods, BioBERT \cite{biobert} was selected as the model used to make relevance predictions in the back-end of the web-based interface of \mname.

\begin{figure}[t]
     \centering
     \begin{subfigure}[t]{1\linewidth}
         \centering
         \captionsetup{justification=centering}
         \includegraphics[width=\linewidth]{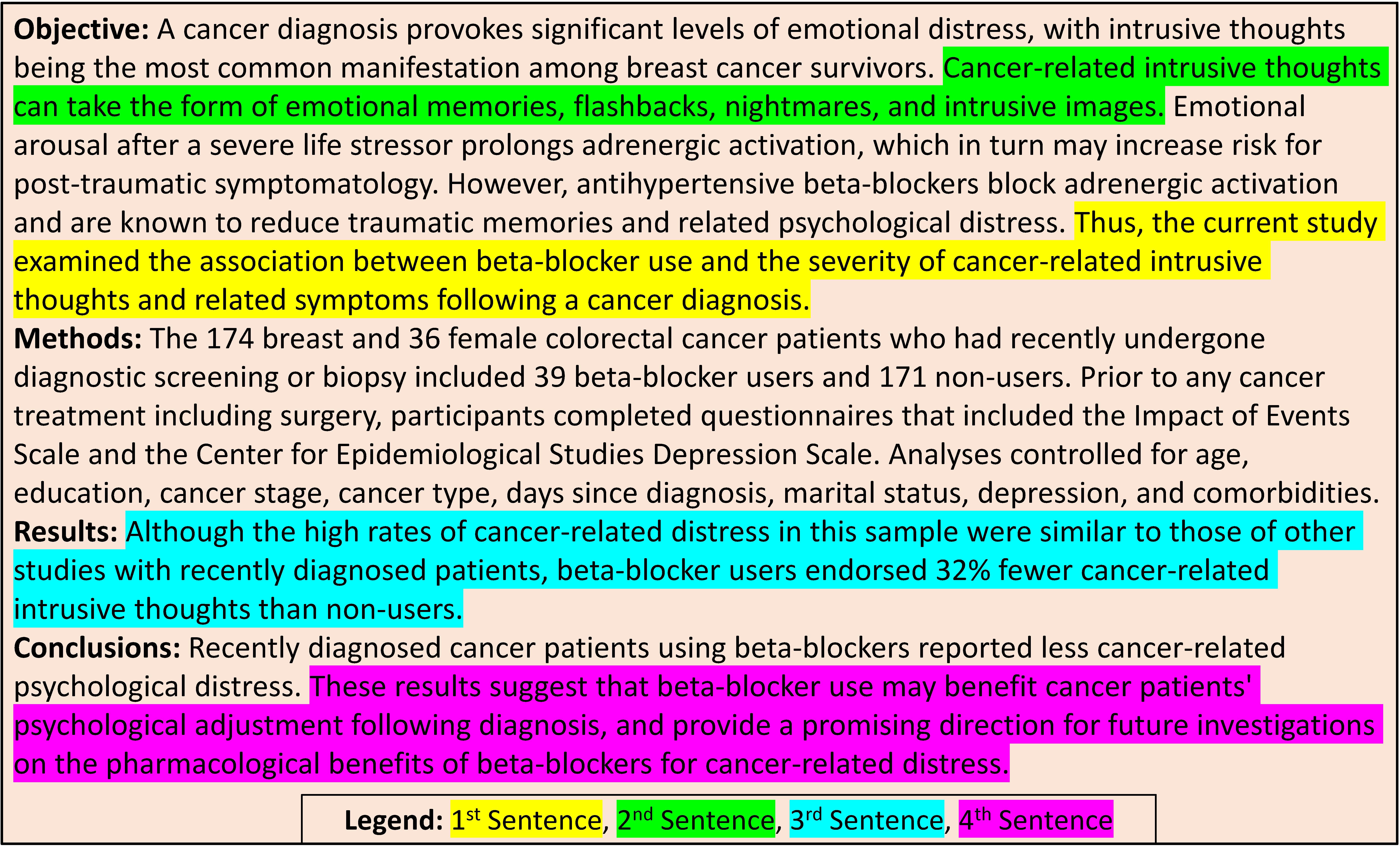}
         \caption{Article with the 4 most relevant sentences according to \relevance highlighted}
         \label{fig:summary_1}
     \end{subfigure}
    %  \hfill
     \begin{subfigure}[t]{1\linewidth}
         \centering
         \captionsetup{justification=centering}
         \includegraphics[width=\linewidth]{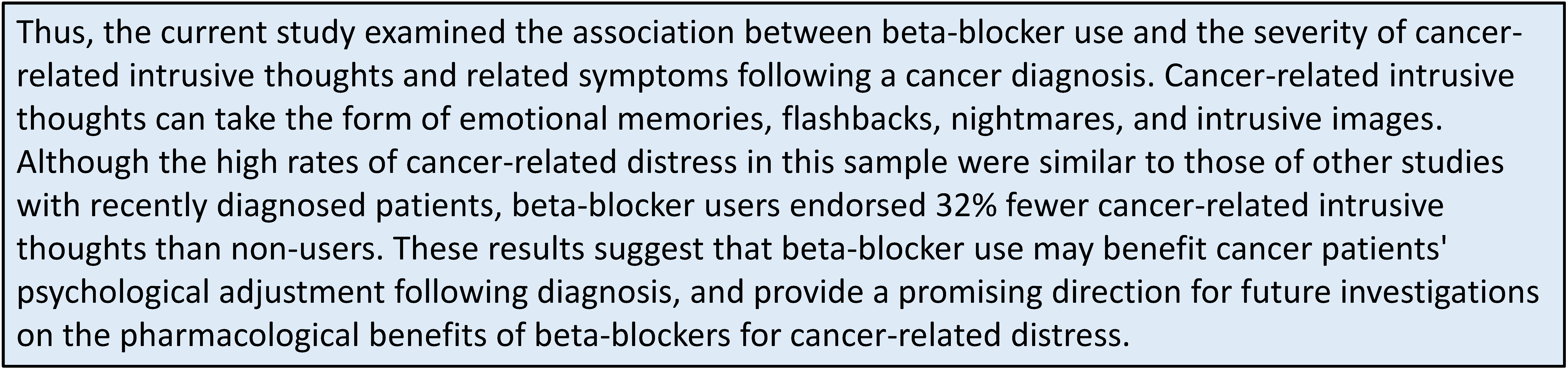}
         \caption{Extractive summary using most relevant sentences}
         \label{fig:summary_2}
     \end{subfigure}
        \caption{Extractive Summarization of PMID 23255459 \cite{pmid_summary_pico_23255459} titled \textit{Beta-blockers may reduce intrusive thoughts in newly diagnosed cancer patients}  by Lindgren et al. with a summary score of 0.588 using \mname}
        \label{fig:summary}
\end{figure}

\vspace{0.5em}

\noindent \textbf{Case Study.} An example of the relevant sentence prediction and subsequent summary formulation using \mname for a PubMed query targeting colorectal cancer articles is demonstrated in Figure \ref{fig:summary}. The article obtains a Summary Score of 0.588 using \summarizer. In this article, titled \textit{Beta-blockers may reduce intrusive thoughts in newly diagnosed cancer patients} by Lindgren et al. \cite{pmid_summary_pico_23255459}, the highest scoring sentence perfectly summarizes the goal of the study. The second sentence provides an example of potential problems faced by the cohort. The third sentence focuses on the study's results, and the fourth concludes the study. The summary score is obtained by averaging the relevance score of each sentence forming the summary. The summary scores of all the articles resulting from the query are shown in Table \ref{tab:summary}. It shows that the model is consistent and obtains a good summary score for all articles, with a maximum score of 0.631 and a minimum score of 0.537.

\section{Patient, Intervention, Outcome Detection}
\label{sec:pio}
% \subsection{Data}
\subsection{Data}
The final component of \mname is aimed at entity recognition of Patient, Intervention, and Outcome in clinical cohort studies. To train \pio for this task, we used the EBM-NLP corpus \cite{pico_data}. The dataset includes 4,970 medical article abstracts with annotations indicating text sequences describing the Participants, Interventions, and Outcome elements of the respective CCS. 4,782 abstracts in the dataset are annotated using crowd-sourced labels. 188 abstracts contain annotations from domain experts with medical training. This test set with gold labels from domain experts is held-out during training and only used to test the performance of the final \pio models.

% Data for PIO extraction was obtained from the EBM-NLP PICO Extraction dataset [22], a corpus
% containing 4,993 abstracts annotated with Participant (P), Intervention (I), and Outcome (O) labels. In
% our framework, we consider Comparator (C) and Intervention (I) to be equal under the same label of
% Intervention (I), as both elements refer to treatments included in a clinical trial.

\subsection{Method}

The pre-trained models are used for \pio: 
\begin{itemize}[leftmargin=*]
    \item BioELECTRA \cite{bioelectra}:  Pre-trained an ELECTRA model on full-text articles from PubMed and PubMed Central.
    \item PubMedBERT \cite{pubmedbert}: Pretrained a BERT  model from scratch using 14 million abstracts from PubMed. 
    \item SciBERT  \cite{scibert}: Pre-trained a BERT model using scientific papers taken from 1.14 million full papers from Semantic Scholar.
    \item BioBERT \cite{biobert}: Initialized using standard BERT \cite{bert} model, and then pre-trained on Biomedical domain texts, which includes PubMed abstracts and PubMed Central full-text articles.
    \item BlueBERT \cite{bluebert}: Initialized using standard BERT \cite{bert} model and pre-trained on PubMed abstracts (4 Billion words) and clinical notes from MIMIC-III (500 Million words) \cite{johnson2016mimic}.
    \item KRISSBERT \cite{krissbert}: Initialized with PubMedBERT \cite{pubmedbert} parameters, and then pre-trained using biomedical entity names from the UMLS ontology \cite{umls} to self-supervise entity linking examples from PubMed abstracts.
    \item SapBERT \cite{sapbert}: Pre-trained a BERT model on the biomedical knowledge graph of UMLS \cite{umls} using self-alignment to cluster synonyms of the same concept.
\end{itemize}

% In addition to the BERT \cite{bert} based models used for relevance prediction described in Section \ref{sec:relevance_method}, we use an additional method for PIO detection named BioELECTRA \cite{bioelectra}. It uses the ELECTRA \cite{clark2020electra} architecture pretrained from scratch on full text articles from PubMed and PubMed Central. This results in the following list of models:    (1) PubMedBERT \cite{pubmedbert}, (2) SapBERT \cite{sapbert}, (3) BlueBERT \cite{bluebert}, (4) KRISSBERT \cite{krissbert}, (5) SciBERT  \cite{scibert}, (6) BioELECTRA, (7) BioBERT \cite{biobert}.

\begin{table*}[t]
\caption{\mname: Participant, Intervention, Outcome Detection Model Performance}
\centering
\resizebox{\linewidth}{!}{%
\begin{tabular}{lcccccccccc}
\toprule
\multirow{2}{*}{\raisebox{-\heavyrulewidth}{Model}} & \multicolumn{3}{|c|}{Precision}    & \multicolumn{3}{|c|}{Recall}       & \multicolumn{3}{|c|}{Micro F1-Score} & Average Micro F1-Score \\ \cmidrule{2-11}
                        & Participant & Intervention & Outcome & Participant & Intervention & Outcome & Participant  & Intervention  & Outcome & P/I/O                    \\ \midrule             
BioELECTRA \cite{bioelectra} & 0.738 & 0.609 & 0.851 & 0.923 & 0.763 & 0.619 & 0.820 & 0.677 & 0.717 & 0.776 \\
PubMedBERT \cite{pubmedbert} & 0.744 & 0.636 & 0.849 & 0.920 & 0.758 & 0.602 & 0.823 & 0.692 & 0.705 & 0.778 \\
SciBERT \cite{scibert}    & 0.743 & 0.609 & 0.854 & 0.910 & 0.750 & 0.607 & 0.818 & 0.673 & 0.710 & 0.773 \\
BioBERT \cite{biobert}   & 0.743 & 0.635 & 0.853 & 0.915 & 0.765 & 0.591 & 0.820 & 0.694 & 0.698 & 0.776 \\
BlueBERT \cite{bluebert}  & 0.724 & 0.635 & 0.852 & 0.916 & 0.749 & 0.593 & 0.809 & 0.687 & 0.700 & 0.771 \\
KRISSBERT \cite{krissbert} & 0.760 & 0.613 & 0.852 & 0.918 & 0.756 & 0.601 & 0.832 & 0.677 & 0.705 & 0.776 \\
SapBERT  \cite{sapbert}  & 0.740 & 0.619 & 0.860 & 0.920 & 0.757 & 0.601 & 0.820 & 0.681 & 0.708 & 0.775 \\ \bottomrule  
\end{tabular}}
\label{tab:pico}
\end{table*}
The labels for each token in the dataset are mapped onto the following 4 labels: Patient, Intervention, Outcome, and None. None represents tokens that are not any of these 3 target PIO entities.

Let $\mathcal{Y^\prime}$ be all the outputs from the model, $\mathcal{Y}$ be all the annotations from the dataset, $\bm{y}^\prime_i$ represent the model prediction, and $\bm{y}_i$ denote the annotation of the $i$-th token. Let $\bm{h}(\mathcal{X})$ represent the output of the transformer architecture. This output, $\bm{h}(\mathcal{X})$, is used as input to a fully-connected layer. So, the output of the $i$-th token is represented by $\bm{y^\prime}_i = \bm{W}^\top \bm{h}(\bm{x}_i) + \bm{b}$. 

% The output of the dense layer, is a vector $\bm{h}(\bm{X}_i) \in \mathbb{R}^{2,496}$. This is followed by a single fully connected layer with softmax activation to predict five different sleep stages:

% \begin{eqnarray*} \label{eq:softmax}
%     \bm{z}_i &=& \bm{W}^\top \bm{h}(\bm{X}_i) + \bm{b} \\
%     \bm{s}_i &=& \text{softmax}(\bm{z}_i)
% \end{eqnarray*}
% where $\bm{W}\in \mathbb{R}^{2,496\times 5}$ is the weight matrix, $\bm{b}\in\mathbb{R}^5$ is the bias vector, and $\bm{s}_i$ is the estimated probabilities of all 5 sleep stages at epoch $i$. 

To train the model, we used cross entropy loss Eq.~\ref{eq:pico_loss}:
\begin{equation} \label{eq:pico_loss}
    L(\bm{y}_i,\; \bm{y^\prime}_i) =  - \sum_{j=1}^4 \bm{y}_i[j] \; log ( \bm{y^\prime}_i[j])
\end{equation} 
where $L(\bm{y}_i,\; \bm{y^\prime}_i)$ is the estimated cross entropy loss for the $i$-th token between annotations $\bm{y} \in \mathbb{R}^4$ and the predicted probabilities $\bm{y^\prime} \in \mathbb{R}^4$, $\bm{y^\prime}_i[j]$ represents the model predictions for the $i$-th token and $j$-th entity.

\vspace{0.5em}
% \subsubsection{Experimental Settings}
\noindent\textbf{Metrics.} The following metrics were used to evaluate the performance of the NER models for PIO detection:
    
% \begin{itemize}[leftmargin=*]
    % \item Accuracy $=\frac{\left|\mathcal{Y}\cap\mathcal{Y}^\prime\right|}{N}$
\begin{equation}
    \begin{aligned}
    \text{Recall}, R^{\left(k\right)}&=\frac{\left|\mathcal{Y}^{\left(k\right)}\cap\mathcal{Y}^{\prime\left(k\right)}\right|}{\left|\mathcal{Y}^{\left(k\right)}\right|}\\
    \text{Precision}, P^{\left(k\right)}&=\frac{\left|\mathcal{Y}^{\left(k\right)}\cap\mathcal{Y}^{\prime\left(k\right)}\right|}{\left|\mathcal{Y}^{\prime\left(k\right)}\right|} \\
 \text{F1 score} &=\frac{2\ \ast\ P\ \ast\ R}{P\ +\ R\ }
    \end{aligned}
\end{equation}
% \end{itemize}

\noindent Given annotations $\mathcal{Y}$, model predictions $\mathcal{Y}^\prime$, $k=$\{Patient, Intervention, Outcome, None\} indicating the entity, $\left|\mathcal{Y}^{\left(k\right)}\right|$ and $\left|{\mathcal{Y}^\prime}^{\left(k\right)}\right|$ represent the number of annotations and model predictions with the label k.
\vspace{0.5em}

\noindent\textbf{Implementation Details.} The \pio was implemented using PyTorch \cite{pytorch,pytorch-lightning} and transformers \cite{huggingface-transformers}. The model was trained using a machine equipped with Intel Xeon Gold 6136 Processor, 376GB RAM, an Nvidia V100 GPU, and CUDA 11.4. A batch size of 6 and a learning rate of $10^{-4}$ were used for training. \pio was trained for 2 epochs using AdamW \cite{adamw} as the optimization method.

The held-out test set was formed using the gold annotated labels in the EBM-NLP corpus \cite{pico_data}. The remaining articles were split randomly in a 9:1 ratio corresponding to the training and validation set and used to optimize training and set hyperparameters. The held-out test set was used to evaluate \pio and compare different baselines.

\subsection{Result}

% \begin{figure}[ht]
%     \centering
%     \captionsetup{justification=centering}
%     \includegraphics[width=\linewidth]{figures/PICO_figure.png}
%     \caption{Participant, Intervention, Outcome Detection of PMID 21453301 \cite{pmid_pico_21453301} titled \textit{Does $\beta$-adrenoceptor blocker therapy improve cancer survival? findings from a population-based retrospective cohort study} using \mname}
%     \label{fig:pio}
% \end{figure}

\begin{figure}[t]
     \centering
     \begin{subfigure}[t]{1\linewidth}
         \centering
         \captionsetup{justification=centering}
         \includegraphics[width=\linewidth]{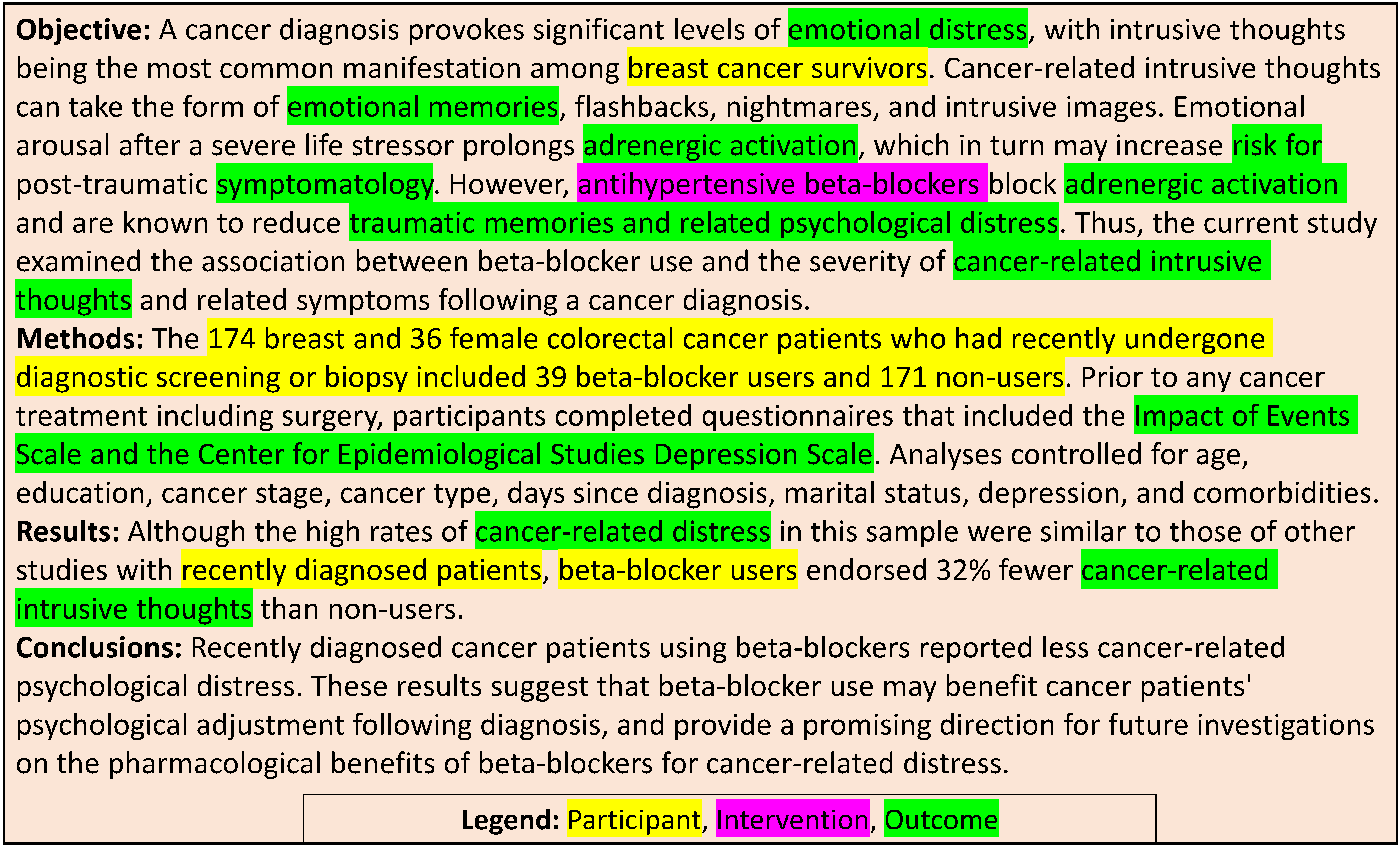}
         \caption{Article with PIO Elements highlighted}
         \label{fig:pico_1}
     \end{subfigure}
    %  \hfill
    
     \begin{subfigure}[t]{1\linewidth}
         \centering
         \captionsetup{justification=centering}
         \includegraphics[width=\linewidth]{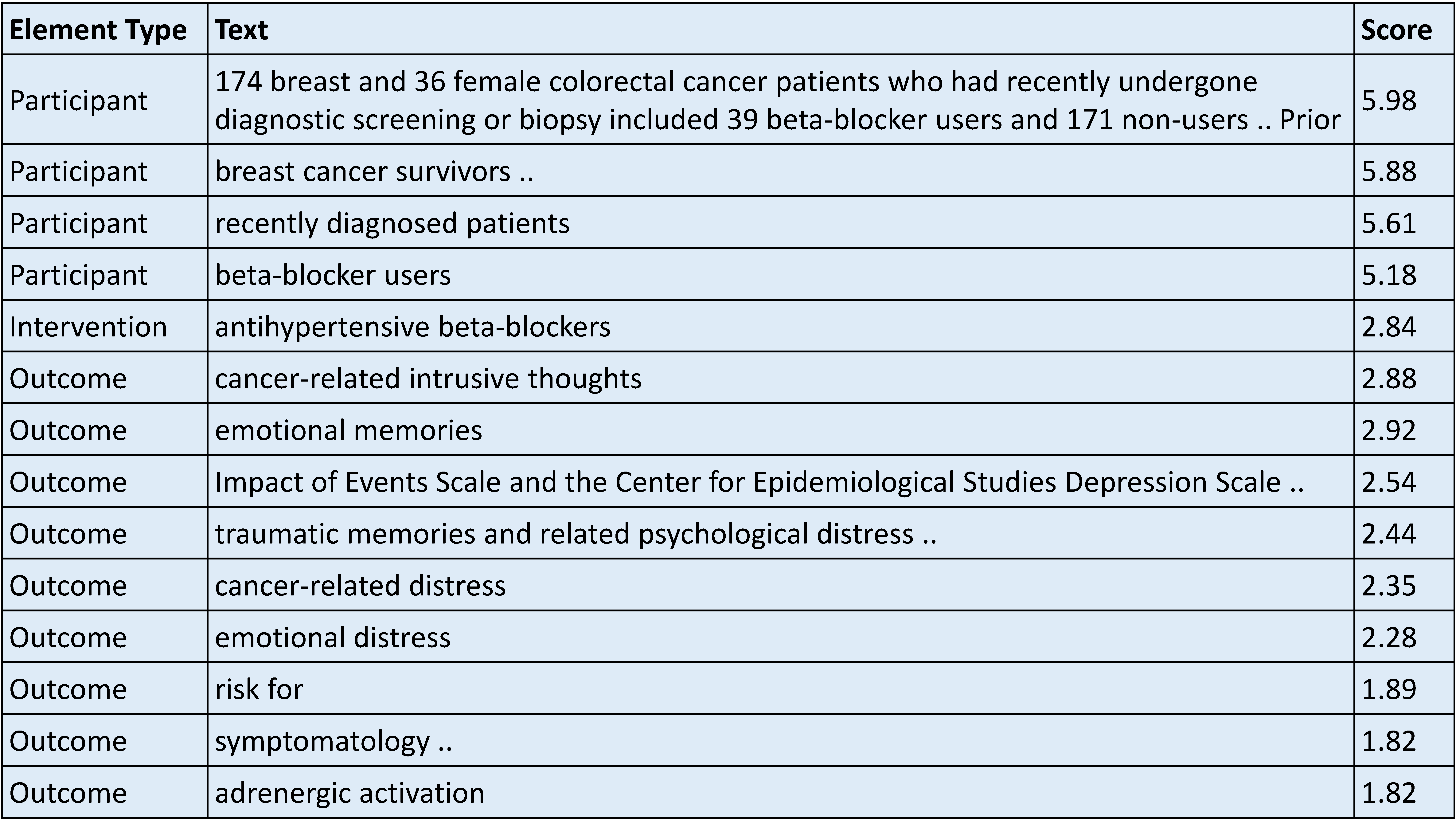}
         \caption{PIO Elements Ranked}
         \label{fig:pico_2}
     \end{subfigure}
        \caption{Participant, Intervention, Outcome (PIO) Detection of PMID 23255459 \cite{pmid_summary_pico_23255459} titled \textit{Beta-blockers may reduce intrusive thoughts in newly diagnosed cancer patients} by Lindgren et al. using \mname}
        \label{fig:pico}
\end{figure}

Table \ref{tab:pico} compares the \pio in \mname using different pre-trained BERT \cite{bert} and ELECTRA \cite{clark2020electra} models. The Average Mirco-F1 $>$77\% shows the efficacy of \pio in detecting the 3 entities: Participants, Intervention, and Outcome. The pre-trained states of these models do not affect the performance after fine-tuning, highlighted by a difference $<1\%$ in the average micro F1-score. \pio is particularly adept in detecting Participants resulting in the highest Recall and Micro-F1 Score among the 3 entities. Due to the highest Average Micro-F1 score among all methods, PubMedBERT \cite{pubmedbert} was selected for the back-end of the web-based interface of \mname.

\vspace{0.5em}

\noindent \textbf{Case Study.} Figure \ref{fig:pico} shows the Participants, Interventions, and Outcomes detected and their respective scores for the same paper expanded upon in Section \ref{sec:relevance_result} for relevance prediction. The paper is titled \textit{Beta-blockers may reduce intrusive thoughts in newly diagnosed cancer patients} by Lindgren et al. \cite{pmid_summary_pico_23255459}. In this paper, participant entities obtain much higher average prediction scores than other entities. The accuracy of participant entity detection across other papers is evident in Table \ref{tab:pico}, where participant entities obtain the highest recall and F1-scores. Overall, the detection of PIO entities across the dataset aligns well with a manual review.

\section{Comparison with Manual Exploration}
The goal of the query used to illustrate the capabilities of \mname is to answer the following question: How do anti-hypertensive drugs impact the outcome of colorectal cancer survival? The advanced PubMed query automatically constructed by \mname shown in Figure 3 is: ((\textquotesingle\textquotesingle colorectal\textquotesingle\textquotesingle\xspace AND (neoplasm OR cancer OR tumor)) OR \textquotesingle\textquotesingle colorectal neoplasms\textquotesingle\textquotesingle\xspace [MeSH]) AND (\textquotesingle\textquotesingle Adrenergic beta-antagonists\textquotesingle\textquotesingle\xspace [MeSH] OR \textquotesingle\textquotesingle Antihypertensive Agents\textquotesingle\textquotesingle\xspace [MeSH] OR \textquotesingle\textquotesingle beta-blockers\textquotesingle\textquotesingle\xspace ) AND (\textquotesingle\textquotesingle Cancer Survivors\textquotesingle\textquotesingle\xspace [MeSH] OR \textquotesingle\textquotesingle cancer survivorship\textquotesingle\textquotesingle\xspace [MeSH] OR \textquotesingle\textquotesingle cancer survivors\textquotesingle\textquotesingle\xspace  OR \textquotesingle\textquotesingle cancer survival\textquotesingle\textquotesingle\xspace ). Appropriate query formatting is critical in finding the most relevant clinical cohort studies. The query mentioned earlier returned 11 studies. A more general PubMed query of \textquotesingle\textquotesingle colorectal cancer\textquotesingle\textquotesingle\xspace at the time of this writing yielded 281,217 studies. A more precise query of \textquotesingle\textquotesingle colorectal cancer AND hypertension\textquotesingle\textquotesingle\xspace  returned 1,617 results. \mname automatically formats the anti-hypertensive drug names and all synonymous versions of the outcome \textquotesingle\textquotesingle cancer survival\textquotesingle\textquotesingle\xspace  to ensure maximal coverage while still restricting the output to the most relevant studies.  

Explicitly comparing \mname to manual exploration by a trained human curator is enlightening. Even if the human curator appropriately formats the advanced PubMed query, there is still substantial time saving with \mname. Here, we compared the exploration time after the selection of relevant articles. Based on timings from trained curator studies \cite{mitchell2015undergraduate}, the average exploration time per relevant article is 29 minutes with a range of 24 to 42 minutes. The variability in manual exploration is based on two factors: the innate skill of the curator and the difficulty of finding the relevant PIO elements in the article based on the article's structure and length. Thus, a  trained curator would take 290 minutes on average to explore only 10 relevant articles compared to the 26.32 seconds required by \mname. 

In addition to time savings, \mname also provides critical context that is not generated during the equivalent manual process. \mname provides the quantitative relevance rankings of each study. The relevance ranking is beneficial for prioritizing the review of large sets of returned relevant articles. The relevance ranking also helps the curator determine how relevant the results of the advanced PubMed query are to the exploratory objective. \mname also generates an extractive summary, which takes in only the most relevant sentences from each study. In the demonstrated example, the extractive summary was constructed using the 4 most relevant sentences. However, the user can easily adjust the number of sentences in each extractive summary. The extractive summary allows for fast and efficient exploration by the human curator. Finally, the automated PIO detection and extraction expedites the formation of study inclusion criteria and preliminary curation steps for a subsequent meta-analysis.

\section{Conclusion}

Recently, there has been an explosion of articles on clinical cohort studies, which are readily available through PubMed. However, the sheer number of articles published daily makes it impossible to read through them to extract relevant information manually. This paper proposes an end-to-end system with a user-friendly graphical interface called \mname, which makes this accessible to anyone. \mname can take a disease as input, generate an advanced query for PubMed, and extract the text from all the resulting articles. Next, it ranks each sentence based on a relevance score, creates an extractive summary of the article along with a summary score, and extracts all Participant, Intervention, and Outcome entities. The \relevance, \summarizer, and \pio are evaluated quantitatively, and case studies are performed to demonstrate their effectiveness. Thus, \mname makes the arduous task of performing large-scale meta-analysis and review feasible by drastically reducing the required time and effort.

\section*{Acknowledgment}
We would like to thank the wonderful team at Morningside Center for Innovative and Affordable Medicine, Emory University for consultation during the study. This research was funded by National Science Foundation CAREER grant 1944247 to C.M, National Institute of Health grant U19-AG056169 sub-award to C.M., and the McCamish Parkinson’s Disease Innovation Program at Georgia Institute of Technology and Emory University to C.M.

% \section*{References}

\bibliographystyle{IEEEtran}
\bibliography{IEEEabrv,refs}

\begin{thebibliography}{10}
\providecommand{\url}[1]{#1}
\csname url@samestyle\endcsname
\providecommand{\newblock}{\relax}
\providecommand{\bibinfo}[2]{#2}
\providecommand{\BIBentrySTDinterwordspacing}{\spaceskip=0pt\relax}
\providecommand{\BIBentryALTinterwordstretchfactor}{4}
\providecommand{\BIBentryALTinterwordspacing}{\spaceskip=\fontdimen2\font plus
\BIBentryALTinterwordstretchfactor\fontdimen3\font minus
  \fontdimen4\font\relax}
\providecommand{\BIBforeignlanguage}[2]{{%
\expandafter\ifx\csname l@#1\endcsname\relax
\typeout{** WARNING: IEEEtran.bst: No hyphenation pattern has been}%
\typeout{** loaded for the language `#1'. Using the pattern for}%
\typeout{** the default language instead.}%
\else
\language=\csname l@#1\endcsname
\fi
#2}}
\providecommand{\BIBdecl}{\relax}
\BIBdecl

\bibitem{white2020pubmed}
J.~White, ``Pubmed 2.0,'' \emph{Medical Reference Services Quarterly}, vol.~39,
  no.~4, pp. 382--387, 2020.

\bibitem{yeo2021alarming}
N.~S.~L. Yeo-Teh and B.~L. Tang, ``An alarming retraction rate for scientific
  publications on coronavirus disease 2019 (covid-19),'' \emph{Accountability
  in research}, vol.~28, no.~1, pp. 47--53, 2021.

\bibitem{mccoy2021biomedical}
K.~McCoy, S.~Gudapati, L.~He, E.~Horlander, D.~Kartchner, S.~Kulkarni,
  N.~Mehra, J.~Prakash, H.~Thenot, S.~V. Vanga \emph{et~al.}, ``Biomedical text
  link prediction for drug discovery: a case study with covid-19,''
  \emph{Pharmaceutics}, vol.~13, no.~6, p. 794, 2021.

\bibitem{burke2007interpretation}
M.~A. Burke and W.~G. Cotts, ``Interpretation of b-type natriuretic peptide in
  cardiac disease and other comorbid conditions,'' \emph{Heart failure
  reviews}, vol.~12, no.~1, pp. 23--36, 2007.

\bibitem{cavailles2013comorbidities}
A.~Cavaill{\`e}s, G.~Brinchault-Rabin, A.~Dixmier, F.~Goupil, C.~Gut-Gobert,
  S.~Marchand-Adam, J.-C. Meurice, H.~Morel, C.~Person-Tacnet, C.~Leroyer
  \emph{et~al.}, ``Comorbidities of copd,'' \emph{European Respiratory Review},
  vol.~22, no. 130, pp. 454--475, 2013.

\bibitem{listerman2011cardiac}
J.~Listerman, V.~Bittner, B.~K. Sanderson, and T.~M. Brown, ``Cardiac
  rehabilitation outcomes: impact of comorbidities and age,'' \emph{Journal of
  cardiopulmonary rehabilitation and prevention}, vol.~31, no.~6, p. 342, 2011.

\bibitem{lang2007non}
C.~C. Lang and D.~M. Mancini, ``Non-cardiac comorbidities in chronic heart
  failure,'' \emph{Heart}, vol.~93, no.~6, pp. 665--671, 2007.

\bibitem{arnold2008age}
M.~Arnold, M.~Halpern, N.~Meier, U.~Fischer, T.~Haefeli, L.~Kappeler,
  C.~Brekenfeld, H.~P. Mattle, and K.~Nedeltchev, ``Age-dependent differences
  in demographics, risk factors, co-morbidity, etiology, management, and
  clinical outcome of acute ischemic stroke,'' \emph{Journal of neurology},
  vol. 255, no.~10, pp. 1503--1507, 2008.

\bibitem{ashwell2012waist}
M.~Ashwell, P.~Gunn, and S.~Gibson, ``Waist-to-height ratio is a better
  screening tool than waist circumference and bmi for adult cardiometabolic
  risk factors: systematic review and meta-analysis,'' \emph{Obesity reviews},
  vol.~13, no.~3, pp. 275--286, 2012.

\bibitem{mohanavelu2021meta}
P.~Mohanavelu, M.~Mutnick, N.~Mehra, B.~White, S.~Kudrimoti,
  K.~Hernandez~Kluesner, X.~Chen, T.~Nguyen, E.~Horlander, H.~Thenot
  \emph{et~al.}, ``Meta-analysis of gastrointestinal adverse events from
  tyrosine kinase inhibitors for chronic myeloid leukemia,'' \emph{Cancers},
  vol.~13, no.~7, p. 1643, 2021.

\bibitem{mitchell2015antecedent}
C.~S. Mitchell, S.~K. Hollinger, S.~D. Goswami, M.~A. Polak, R.~H. Lee, and
  J.~D. Glass, ``Antecedent disease is less prevalent in amyotrophic lateral
  sclerosis,'' \emph{Neurodegenerative Diseases}, vol.~15, no.~2, pp. 109--113,
  2015.

\bibitem{makhoul2020epidemiological}
M.~Makhoul, H.~H. Ayoub, H.~Chemaitelly, S.~Seedat, G.~R. Mumtaz, S.~Al-Omari,
  and L.~J. Abu-Raddad, ``Epidemiological impact of sars-cov-2 vaccination:
  Mathematical modeling analyses,'' \emph{Vaccines}, vol.~8, no.~4, p. 668,
  2020.

\bibitem{pritchard2021impact}
E.~Pritchard, P.~C. Matthews, N.~Stoesser, D.~W. Eyre, O.~Gethings, K.-D.
  Vihta, J.~Jones, T.~House, H.~VanSteenHouse, I.~Bell \emph{et~al.}, ``Impact
  of vaccination on new sars-cov-2 infections in the united kingdom,''
  \emph{Nature medicine}, vol.~27, no.~8, pp. 1370--1378, 2021.

\bibitem{shattock2022impact}
A.~J. Shattock, E.~A. Le~Rutte, R.~P. D{\"u}nner, S.~Sen, S.~L. Kelly,
  N.~Chitnis, and M.~A. Penny, ``Impact of vaccination and non-pharmaceutical
  interventions on sars-cov-2 dynamics in switzerland,'' \emph{Epidemics},
  vol.~38, p. 100535, 2022.

\bibitem{mitchell2015undergraduate}
C.~S. Mitchell, A.~Cates, R.~B. Kim, and S.~K. Hollinger, ``Undergraduate
  biocuration: developing tomorrow’s researchers while mining today’s
  data,'' \emph{Journal of Undergraduate Neuroscience Education}, vol.~14,
  no.~1, p. A56, 2015.

\bibitem{tsafnat2013automation}
G.~Tsafnat, A.~Dunn, P.~Glasziou, and E.~Coiera, ``The automation of systematic
  reviews,'' 2013.

\bibitem{sentiment}
\BIBentryALTinterwordspacing
C.~Du, H.~Sun, J.~Wang, Q.~Qi, and J.~Liao, ``Adversarial and domain-aware
  {BERT} for cross-domain sentiment analysis,'' in \emph{Proceedings of the
  58th Annual Meeting of the Association for Computational Linguistics}.\hskip
  1em plus 0.5em minus 0.4em\relax Online: Association for Computational
  Linguistics, Jul. 2020, pp. 4019--4028. [Online]. Available:
  \url{https://aclanthology.org/2020.acl-main.370}
\BIBentrySTDinterwordspacing

\bibitem{translation1}
\BIBentryALTinterwordspacing
K.~Chen, R.~Wang, M.~Utiyama, and E.~Sumita, ``Neural machine translation with
  reordering embeddings,'' in \emph{Proceedings of the 57th Annual Meeting of
  the Association for Computational Linguistics}.\hskip 1em plus 0.5em minus
  0.4em\relax Florence, Italy: Association for Computational Linguistics, Jul.
  2019, pp. 1787--1799. [Online]. Available:
  \url{https://aclanthology.org/P19-1174}
\BIBentrySTDinterwordspacing

\bibitem{translation2}
T.~Nishihara, A.~Tamura, T.~Ninomiya, Y.~Omote, and H.~Nakayama, ``Supervised
  visual attention for multimodal neural machine translation,'' in
  \emph{Proceedings of the 28th International Conference on Computational
  Linguistics}, 2020, pp. 4304--4314.

\bibitem{anyanwu2005semrank}
K.~Anyanwu, A.~Maduko, and A.~Sheth, ``Semrank: ranking complex relationship
  search results on the semantic web,'' in \emph{Proceedings of the 14th
  international conference on World Wide Web}, 2005, pp. 117--127.

\bibitem{belter2017relevance}
C.~W. Belter, ``A relevance ranking method for citation-based search results,''
  \emph{Scientometrics}, vol. 112, no.~2, pp. 731--746, 2017.

\bibitem{gao2020toward}
R.~Gao and C.~Shah, ``Toward creating a fairer ranking in search engine
  results,'' \emph{Information Processing \& Management}, vol.~57, no.~1, p.
  102138, 2020.

\bibitem{zhang2020pegasus}
J.~Zhang, Y.~Zhao, M.~Saleh, and P.~Liu, ``Pegasus: Pre-training with extracted
  gap-sentences for abstractive summarization,'' in \emph{International
  Conference on Machine Learning}.\hskip 1em plus 0.5em minus 0.4em\relax PMLR,
  2020, pp. 11\,328--11\,339.

\bibitem{maynez2020faithfulness}
J.~Maynez, S.~Narayan, B.~Bohnet, and R.~McDonald, ``On faithfulness and
  factuality in abstractive summarization,'' \emph{arXiv preprint
  arXiv:2005.00661}, 2020.

\bibitem{gehrmann2018bottom}
S.~Gehrmann, Y.~Deng, and A.~M. Rush, ``Bottom-up abstractive summarization,''
  \emph{arXiv preprint arXiv:1808.10792}, 2018.

\bibitem{zhong2020extractive}
M.~Zhong, P.~Liu, Y.~Chen, D.~Wang, X.~Qiu, and X.~Huang, ``Extractive
  summarization as text matching,'' \emph{arXiv preprint arXiv:2004.08795},
  2020.

\bibitem{liu2019fine}
Y.~Liu, ``Fine-tune bert for extractive summarization,'' \emph{arXiv preprint
  arXiv:1903.10318}, 2019.

\bibitem{bert}
J.~Devlin, M.-W. Chang, K.~Lee, and K.~Toutanova, ``Bert: Pre-training of deep
  bidirectional transformers for language understanding,'' \emph{arXiv preprint
  arXiv:1810.04805}, 2018.

\bibitem{clark2020electra}
K.~Clark, M.-T. Luong, Q.~V. Le, and C.~D. Manning, ``Electra: Pre-training
  text encoders as discriminators rather than generators,'' \emph{arXiv
  preprint arXiv:2003.10555}, 2020.

\bibitem{streamlit}
\BIBentryALTinterwordspacing
``Streamlit: the fastest way to build and share data apps.'' [Online].
  Available: \url{https://streamlit.io/}
\BIBentrySTDinterwordspacing

\bibitem{biopython_1}
P.~J. Cock, T.~Antao, J.~T. Chang, B.~A. Chapman, C.~J. Cox, A.~Dalke,
  I.~Friedberg, T.~Hamelryck, F.~Kauff, B.~Wilczynski \emph{et~al.},
  ``Biopython: freely available python tools for computational molecular
  biology and bioinformatics,'' \emph{Bioinformatics}, vol.~25, no.~11, pp.
  1422--1423, 2009.

\bibitem{biopython_2}
B.~Chapman and J.~Chang, ``Biopython: Python tools for computational biology,''
  \emph{ACM Sigbio Newsletter}, vol.~20, no.~2, pp. 15--19, 2000.

\bibitem{scispacy}
\BIBentryALTinterwordspacing
M.~Neumann, D.~King, I.~Beltagy, and W.~Ammar, ``{S}cispa{C}y: {F}ast and
  {R}obust {M}odels for {B}iomedical {N}atural {L}anguage {P}rocessing,'' in
  \emph{Proceedings of the 18th BioNLP Workshop and Shared Task}.\hskip 1em
  plus 0.5em minus 0.4em\relax Florence, Italy: Association for Computational
  Linguistics, Aug. 2019, pp. 319--327. [Online]. Available:
  \url{https://www.aclweb.org/anthology/W19-5034}
\BIBentrySTDinterwordspacing

\bibitem{relevance_data1}
E.~Lehman, J.~DeYoung, R.~Barzilay, and B.~C. Wallace, ``Inferring which
  medical treatments work from reports of clinical trials,'' in
  \emph{Proceedings of the North American Chapter of the Association for
  Computational Linguistics (NAACL)}, 2019, pp. 3705--3717.

\bibitem{relevance_data2}
J.~DeYoung, E.~Lehman, B.~Nye, I.~J. Marshall, and B.~C. Wallace, ``Evidence
  inference 2.0: More data, better models,'' 2020.

\bibitem{relevance_model}
B.~E. Nye, J.~DeYoung, E.~Lehman, A.~Nenkova, I.~J. Marshall, and B.~C.
  Wallace, ``Understanding clinical trial reports: Extracting medical entities
  and their relations,'' \emph{AMIA Summits on Translational Science
  Proceedings}, vol. 2021, p. 485, 2021.

\bibitem{biobert}
J.~Lee, W.~Yoon, S.~Kim, D.~Kim, S.~Kim, C.~H. So, and J.~Kang, ``Biobert: a
  pre-trained biomedical language representation model for biomedical text
  mining,'' \emph{Bioinformatics}, vol.~36, no.~4, pp. 1234--1240, 2020.

\bibitem{pubmedbert}
Y.~Gu, R.~Tinn, H.~Cheng, M.~Lucas, N.~Usuyama, X.~Liu, T.~Naumann, J.~Gao, and
  H.~Poon, ``Domain-specific language model pretraining for biomedical natural
  language processing,'' 2020.

\bibitem{sapbert}
\BIBentryALTinterwordspacing
F.~Liu, E.~Shareghi, Z.~Meng, M.~Basaldella, and N.~Collier, ``Self-alignment
  pretraining for biomedical entity representations,'' in \emph{Proceedings of
  the 2021 Conference of the North American Chapter of the Association for
  Computational Linguistics: Human Language Technologies}.\hskip 1em plus 0.5em
  minus 0.4em\relax Online: Association for Computational Linguistics, Jun.
  2021, pp. 4228--4238. [Online]. Available:
  \url{https://www.aclweb.org/anthology/2021.naacl-main.334}
\BIBentrySTDinterwordspacing

\bibitem{umls}
O.~Bodenreider, ``The unified medical language system (umls): integrating
  biomedical terminology,'' \emph{Nucleic acids research}, vol.~32, no.
  suppl\_1, pp. D267--D270, 2004.

\bibitem{bluebert}
Y.~Peng, S.~Yan, and Z.~Lu, ``Transfer learning in biomedical natural language
  processing: An evaluation of bert and elmo on ten benchmarking datasets,'' in
  \emph{Proceedings of the 2019 Workshop on Biomedical Natural Language
  Processing (BioNLP 2019)}, 2019, pp. 58--65.

\bibitem{johnson2016mimic}
A.~E. Johnson, T.~J. Pollard, L.~Shen, L.-w.~H. Lehman, M.~Feng, M.~Ghassemi,
  B.~Moody, P.~Szolovits, L.~Anthony~Celi, and R.~G. Mark, ``Mimic-iii, a
  freely accessible critical care database,'' \emph{Scientific data}, vol.~3,
  no.~1, pp. 1--9, 2016.

\bibitem{krissbert}
S.~Zhang, H.~Cheng, S.~Vashishth, C.~Wong, J.~Xiao, X.~Liu, T.~Naumann, J.~Gao,
  and H.~Poon, ``Knowledge-rich self-supervised entity linking,'' \emph{arXiv
  preprint arXiv:2112.07887}, 2021.

\bibitem{scibert}
\BIBentryALTinterwordspacing
I.~Beltagy, K.~Lo, and A.~Cohan, ``{S}ci{BERT}: A pretrained language model for
  scientific text,'' in \emph{Proceedings of the 2019 Conference on Empirical
  Methods in Natural Language Processing and the 9th International Joint
  Conference on Natural Language Processing (EMNLP-IJCNLP)}.\hskip 1em plus
  0.5em minus 0.4em\relax Hong Kong, China: Association for Computational
  Linguistics, Nov. 2019, pp. 3615--3620. [Online]. Available:
  \url{https://aclanthology.org/D19-1371}
\BIBentrySTDinterwordspacing

\bibitem{pytorch}
\BIBentryALTinterwordspacing
A.~Paszke, S.~Gross, F.~Massa, A.~Lerer, J.~Bradbury, G.~Chanan, T.~Killeen,
  Z.~Lin, N.~Gimelshein, L.~Antiga, A.~Desmaison, A.~Kopf, E.~Yang, Z.~DeVito,
  M.~Raison, A.~Tejani, S.~Chilamkurthy, B.~Steiner, L.~Fang, J.~Bai, and
  S.~Chintala, ``Pytorch: An imperative style, high-performance deep learning
  library,'' in \emph{Advances in Neural Information Processing Systems 32},
  H.~Wallach, H.~Larochelle, A.~Beygelzimer, F.~d\textquotesingle
  Alch\'{e}-Buc, E.~Fox, and R.~Garnett, Eds.\hskip 1em plus 0.5em minus
  0.4em\relax Curran Associates, Inc., 2019, pp. 8024--8035. [Online].
  Available:
  \url{http://papers.neurips.cc/paper/9015-pytorch-an-imperative-style-high-performance-deep-learning-library.pdf}
\BIBentrySTDinterwordspacing

\bibitem{pytorch-lightning}
W.~Falcon \emph{et~al.}, ``Pytorch lightning,'' \emph{GitHub. Note:
  https://github. com/PyTorchLightning/pytorch-lightning}, vol.~3, no.~6, 2019.

\bibitem{huggingface-transformers}
\BIBentryALTinterwordspacing
T.~Wolf, L.~Debut, V.~Sanh, J.~Chaumond, C.~Delangue, A.~Moi, P.~Cistac,
  T.~Rault, R.~Louf, M.~Funtowicz, J.~Davison, S.~Shleifer, P.~von Platen,
  C.~Ma, Y.~Jernite, J.~Plu, C.~Xu, T.~L. Scao, S.~Gugger, M.~Drame, Q.~Lhoest,
  and A.~M. Rush, ``Transformers: State-of-the-art natural language
  processing,'' in \emph{Proceedings of the 2020 Conference on Empirical
  Methods in Natural Language Processing: System Demonstrations}.\hskip 1em
  plus 0.5em minus 0.4em\relax Online: Association for Computational
  Linguistics, Oct. 2020, pp. 38--45. [Online]. Available:
  \url{https://www.aclweb.org/anthology/2020.emnlp-demos.6}
\BIBentrySTDinterwordspacing

\bibitem{adam}
D.~P. Kingma and J.~Ba, ``Adam: A method for stochastic optimization,''
  \emph{arXiv preprint arXiv:1412.6980}, 2014.

\bibitem{24050955}
B.~Hicks, L.~Murray, D.~Powe, C.~Hughes, and C.~Cardwell, ``$\beta$-blocker
  usage and colorectal cancer mortality: a nested case--control study in the uk
  clinical practice research datalink cohort,'' \emph{Annals of oncology},
  vol.~24, no.~12, pp. 3100--3106, 2013.

\bibitem{35881046}
R.~Udumyan, E.~Botteri, T.~Jerlstrom, S.~Montgomery, K.~E. Smedby, and K.~Fall,
  ``Beta-blocker use and urothelial bladder cancer survival: a swedish
  register-based cohort study,'' \emph{Acta Oncologica}, pp. 1--9, 2022.

\bibitem{29858097}
R.~P. Musselman, S.~Bennett, W.~Li, M.~Mamdani, T.~Gomes, C.~van Walraven,
  R.~Boushey, O.~Al-Obeed, M.~Al-Omran, and R.~C. Auer, ``Association between
  perioperative beta blocker use and cancer survival following surgical
  resection,'' \emph{European Journal of Surgical Oncology}, vol.~44, no.~8,
  pp. 1164--1169, 2018.

\bibitem{29846174}
F.~Peng, D.~Hu, X.~Lin, B.~Liang, Y.~Chen, H.~Zhang, Y.~Xia, J.~Lin, X.~Zheng,
  and W.~Niu, ``Impact of long-term antihypertensive and antidiabetic
  medications on the prognosis of post-surgical colorectal cancer: the fujian
  prospective investigation of cancer (fiesta) study,'' \emph{Aging (Albany
  NY)}, vol.~10, no.~5, p. 1166, 2018.

\bibitem{34843550}
J.~G. Trogdon, K.~Amin, P.~Gupta, B.~Y. Urick, K.~E. Reeder-Hayes, J.~F.
  Farley, S.~B. Wheeler, L.~Spees, and J.~L. Lund, ``Providers’ mediating
  role for medication adherence among cancer survivors,'' \emph{PloS one},
  vol.~16, no.~11, p. e0260358, 2021.

\bibitem{31062847}
Y.~Cui, W.~Wen, T.~Zheng, H.~Li, Y.-T. Gao, H.~Cai, M.~You, J.~Gao, G.~Yang,
  W.~Zheng \emph{et~al.}, ``Use of antihypertensive medications and survival
  rates for breast, colorectal, lung, or stomach cancer,'' \emph{American
  Journal of Epidemiology}, vol. 188, no.~8, pp. 1512--1528, 2019.

\bibitem{35725814}
L.~L. L{\o}fling, N.~C. St{\o}er, E.~K. Sloan, A.~Chang, S.~Gandini, G.~Ursin,
  and E.~Botteri, ``$\beta$-blockers and breast cancer survival by molecular
  subtypes: a population-based cohort study and meta-analysis,'' \emph{British
  Journal of Cancer}, pp. 1--11, 2022.

\bibitem{pmid_summary_pico_23255459}
M.~E. Lindgren, C.~P. Fagundes, C.~M. Alfano, S.~P. Povoski, D.~M. Agnese,
  M.~W. Arnold, W.~B. Farrar, L.~D. Yee, W.~E. Carson, C.~R. Schmidt
  \emph{et~al.}, ``Beta-blockers may reduce intrusive thoughts in newly
  diagnosed cancer patients,'' \emph{Psycho-oncology}, vol.~22, no.~8, pp.
  1889--1894, 2013.

\bibitem{30917783}
E.~J. Bowles, O.~Yu, R.~Ziebell, L.~Chen, D.~M. Boudreau, D.~P. Ritzwoller,
  R.~A. Hubbard, J.~M. Boggs, A.~N. Burnett-Hartman, A.~Sterrett \emph{et~al.},
  ``Cardiovascular medication use and risks of colon cancer recurrences and
  additional cancer events: a cohort study,'' \emph{BMC cancer}, vol.~19,
  no.~1, pp. 1--12, 2019.

\bibitem{21453301}
S.~M. Shah, I.~M. Carey, C.~G. Owen, T.~Harris, S.~DeWilde, and D.~G. Cook,
  ``Does $\beta$-adrenoceptor blocker therapy improve cancer survival? findings
  from a population-based retrospective cohort study,'' \emph{British journal
  of clinical pharmacology}, vol.~72, no.~1, pp. 157--161, 2011.

\bibitem{lavie2004significance}
A.~Lavie, K.~Sagae, and S.~Jayaraman, ``The significance of recall in automatic
  metrics for mt evaluation,'' in \emph{Conference of the Association for
  Machine Translation in the Americas}.\hskip 1em plus 0.5em minus 0.4em\relax
  Springer, 2004, pp. 134--143.

\bibitem{su1994relevance}
L.~T. Su, ``The relevance of recall and precision in user evaluation,''
  \emph{Journal of the American Society for Information Science}, vol.~45,
  no.~3, pp. 207--217, 1994.

\bibitem{nenkova2006summarization}
A.~Nenkova, ``Summarization evaluation for text and speech: issues and
  approaches,'' in \emph{Ninth International Conference on Spoken Language
  Processing}, 2006.

\bibitem{pico_data}
\BIBentryALTinterwordspacing
B.~Nye, J.~J. Li, R.~Patel, Y.~Yang, I.~Marshall, A.~Nenkova, and B.~Wallace,
  ``A corpus with multi-level annotations of patients, interventions and
  outcomes to support language processing for medical literature,'' in
  \emph{Proceedings of the 56th Annual Meeting of the Association for
  Computational Linguistics (Volume 1: Long Papers)}.\hskip 1em plus 0.5em
  minus 0.4em\relax Melbourne, Australia: Association for Computational
  Linguistics, Jul. 2018, pp. 197--207. [Online]. Available:
  \url{https://aclanthology.org/P18-1019}
\BIBentrySTDinterwordspacing

\bibitem{bioelectra}
\BIBentryALTinterwordspacing
K.~r. Kanakarajan, B.~Kundumani, and M.~Sankarasubbu,
  ``{B}io{ELECTRA}:pretrained biomedical text encoder using discriminators,''
  in \emph{Proceedings of the 20th Workshop on Biomedical Language
  Processing}.\hskip 1em plus 0.5em minus 0.4em\relax Online: Association for
  Computational Linguistics, Jun. 2021, pp. 143--154. [Online]. Available:
  \url{https://aclanthology.org/2021.bionlp-1.16}
\BIBentrySTDinterwordspacing

\bibitem{adamw}
I.~Loshchilov and F.~Hutter, ``Decoupled weight decay regularization,''
  \emph{arXiv preprint arXiv:1711.05101}, 2017.

\end{thebibliography}

\end{document}